\documentclass[11pt]{article}

\usepackage[final]{acl}

\usepackage{times}
\usepackage{latexsym}

\usepackage[T2A,T1]{fontenc}

\usepackage[utf8]{inputenc}

\usepackage{microtype}

\usepackage{inconsolata}

\usepackage{graphicx}

\usepackage[russian,english]{babel}
\usepackage{color, colortbl}
\usepackage{xcolor}

\definecolor{LightCyan}{rgb}{0.88,1,1}
\definecolor{lemonchiffon}{rgb}{1.0, 0.98, 0.8}

\usepackage{multirow}
\usepackage{float}

\title{Bukva: Russian Sign Language Alphabet}

\author{
  Petr Surovtsev\textsuperscript{1} \\
  \texttt{pysurovtsev@daimld.tech}
  \And
  Alexander Nagaev\textsuperscript{1} \\
  \texttt{aonagaev@daimld.tech}
  \And
  Karina Kvanchiani\textsuperscript{1} \\
  \texttt{karinakvanciani@gmail.com}
  \AND
  Elizaveta Petrova\textsuperscript{1} \\
  \texttt{kleinsbotle@gmail.com} \\
  \And
  Alexander Kapitanov\textsuperscript{1} \\
  \texttt{aakapitanov@daimld.tech} \\
\AND
\textsuperscript{1}\textnormal{Department of Artificial Intelligence and Machine Learning Development, DAIMLD}
}

\begin{document}
\maketitle

\begin{abstract}
\label{abstract}

This paper investigates the recognition of the Russian fingerspelling alphabet, also known as the Russian Sign Language (RSL) dactyl. Dactyl is a component of sign languages where distinct hand movements represent individual letters of a written language. This method is used to spell words without specific signs, such as proper nouns or technical terms. An alphabet learning simulator is one meaningful application of isolated dactyl recognition. There is a notable issue of data shortage in isolated dactyl recognition: existing Russian dactyl datasets lack subject heterogeneity, contain insufficient samples, or cover only static signs. We provide Bukva, the first open-source video dataset for RSL dactyl recognition that is free of all these shortcomings. It contains 3,757 videos with more than 100 samples for each RSL alphabet sign, including dynamic ones. We utilized crowdsourcing platforms to increase the subjects' heterogeneity, resulting in the participation of 155 deaf and hard-of-hearing experts in the dataset creation. We use a TSM (Temporal Shift Module) block to handle static and dynamic signs effectively, achieving 83.6\% top-1 accuracy with real-time inference on CPU only. The dataset, demo code, and pre-trained models will be made publicly available\footnote{\url{https://github.com/ai-forever/bukva}}.

\end{abstract}

\section{Introduction}
\label{sec:intro}

According to the World Health Organization, over 430 million people worldwide have disabling hearing loss\footnote{\url{https://www.who.int/news-room/fact-sheets/detail/deafness-and-hearing-loss}}, and many deaf and hard-of-hearing people use sign language. According to the statistical report for 2019, there are more than 150 thousand deaf sign language speakers in Russia\footnote{\url{https://voginfo.ru/about/}}. Such languages are unique in terms of rules and grammar in different countries. A fundamental component of sign language is the SL alphabet called dactyl. It visually represents the words of the spoken language letter by letter, a process known as fingerspelling or dactylology. Fingerspelling spells words without a distinct sign, such as proper names and specialized terms, e.g., technical and scientific words and names of people or animals. Also, any word can be fingerspelled letter by letter.

Fingerspelling recognition is an urgent task with vast potential for practical applications. Such a system can be implemented at train stations and airports to indicate toponyms and in hospitals when describing diagnoses and medications. Schools, universities, and other institutions also require a dactylology recognition system to simplify communication using special terms. Another helpful application case is the creation of an interactive SL trainer. The user is shown an example of performing a sign, after which the recognition system checks whether the user repeated it correctly. Since there is a lack of sign language teachers and free courses with feedback, such trainers could improve accessibility to learning Russian Sign Language (RSL). Throughout this work, we consulted with members of the ``All-Russian Society of the Deaf'' (VOG), who confirmed that a dactyl simulator could simplify learning the alphabet for further usage in fingerspelling. A similar application exists for American Sign Language --- the browser-based fingerspelling.xyz\footnote{\url{https://fingerspelling.xyz}}. While fingerspelling applications involve continuous dactyl recognition, the SL trainer represents an isolated task since using a fingerspelling model would be redundant. This paper focuses on recognizing isolated letters to build the dactyl recognition system for a sign language alphabet trainer. It can also serve as a baseline for continuous fingerspelling recognition. 

Moreover, fingerspelling is a key bottleneck for sign language translation (SLT): proper nouns, named entities, and out-of-vocabulary technical terms are conveyed almost entirely through the alphabet, and these are precisely the tokens that most often break gloss-free SLT in low-resource settings. Robust isolated-letter recognition is therefore a building block for downstream continuous SLT, not only an educational tool.

Some unique challenges are presented in the dactyl recognition task. It involves smaller, faster movements that are more difficult to recognize than regular sign language. Since dactyl movements are short and quick, motion blur can have a significant negative impact, requiring the network to be more robust to alterations in data quality. The presence of static and dynamic letters in numerous dactyl alphabets makes it impossible to create a dataset containing only images. Moreover, some dactyl letters in RSL can appear similarly, making recognition problematic. For instance, \foreignlanguage{russian}{``л''} (``l'') -- \foreignlanguage{russian}{``п''} (``p'') and \foreignlanguage{russian}{``м''} (``m'') -- \foreignlanguage{russian}{``т''} (``t'') have comparable configurations and are distinguished only by the spacing between fingers. Similarly, \foreignlanguage{russian}{``ё''} (``yo'') differs from \foreignlanguage{russian}{``е''} (``e'') solely in the hand movement.

Given the use cases discussed above, it is clear that the dactyl recognition system must be robust to variations in the background, lighting, and subjects. It must also operate in real-time and handle videos of varying quality and resolution. Striving for high quality is also crucial, as even a small error can have significant consequences. These requirements place certain constraints on the model: it must be efficient enough for real-time processing and of acceptable size to avoid overfitting caused by the relatively small number of classes in the dactyl alphabet. Furthermore, the task's specifics restrict the training data: the dataset must be heterogeneous across subjects and scenes, balanced across classes, and contain sufficient samples. Existing Russian dactyl datasets are unsuitable because they lack diverse subjects and scenes and sufficient samples. Besides, some cannot be used to build a full-fledged SL alphabet recognition system as they only cover static signs, omitting dynamic ones.

This work presents Bukva, the first full-fledged open-source video dataset for classifying the RSL alphabet. By full-fledged, we mean a dataset that simultaneously (1) covers all 33 RSL alphabet letters, including dynamic ones, (2) provides a sufficient number of samples per sign (more than 100), and (3) features heterogeneous signers, scenes, and recording conditions; none of the existing RSL alphabet datasets, including the open-source Slovo \citep{slovo}, satisfies all these criteria. RSL remains an underrepresented sign language in terms of available resources. Since RSL significantly differs from the Russian language, we regularly consult with RSL teachers and native speakers to create the data, considering the language's rules and nuances. The resulting dataset contains 3,757 high-quality videos divided into 33 classes. The videos were captured using various devices and featured 155 RSL speakers as signers, showcasing diverse scenes and lighting conditions. We integrated the RSL exam into the data creation pipeline to limit video recording access to individuals without RSL knowledge. Besides, all videos were filtered and validated to ensure the quality of the data. Bukva also includes an extra ``no sign'' class to assist the model in determining the sign frames from the entire video.

In experiments, we employ the TSM block \citep{lin2019tsm} to balance different frame variability according to the temporal component for static and dynamic signs. It enables the 2D convolutions to process video data without increasing the number of parameters. The experiments demonstrate that a model trained using the proposed dataset achieves high metrics, with a top-1 accuracy of 83.6\% and real-time CPU performance. Furthermore, we also provide a demo application illustrating the operation of a simulator for teaching dactyl in Russian sign language. The Bukva dataset, demo code, and pre-trained models will be made publicly available.

Our contributions can be summarized as follows: (1) Bukva involves the largest number and diversity of signers among RSL alphabet datasets (155); (2) all samples are HD or FullHD videos; (3) the dataset covers all 33 RSL alphabet letters, including dynamic ones; (4) the dataset, demo code, and pre-trained models are open-sourced; (5) we describe a reproducible crowd-based collection and annotation methodology --- proficiency-gated recruitment, multi-annotator validation, and sign-interval annotation --- transferable to other low-resource sign languages

\begin{figure}[t]
  \centering
  \includegraphics[width=0.9\linewidth]{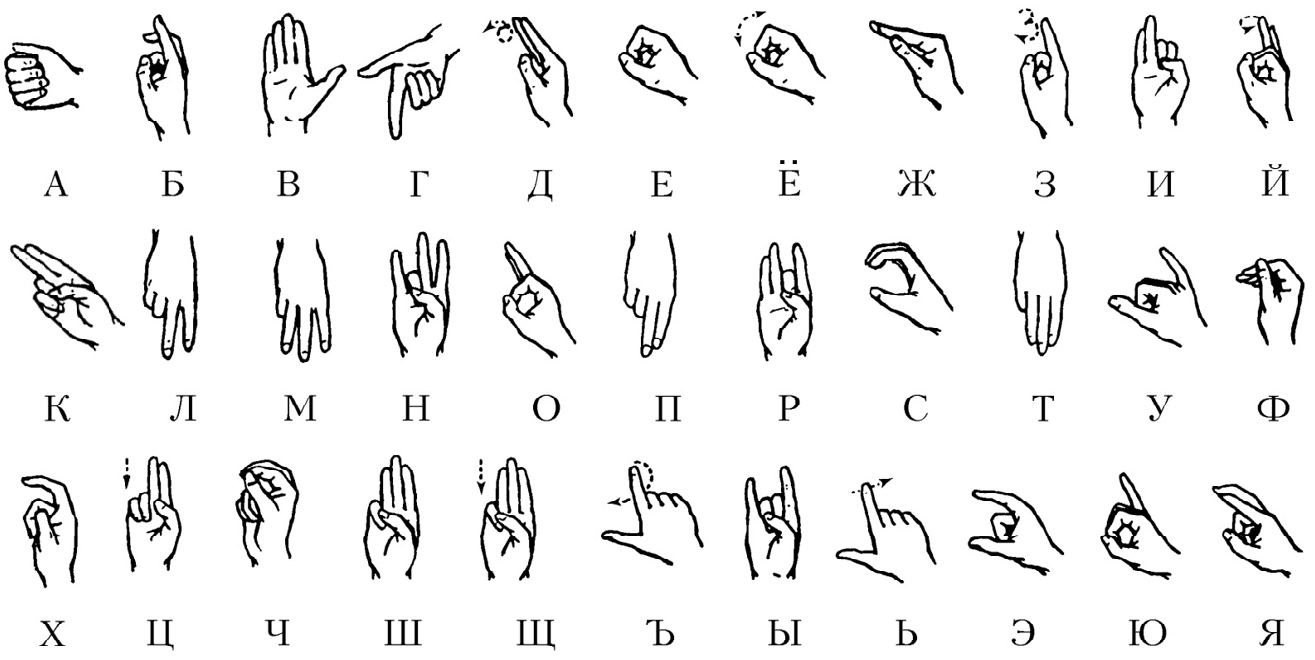}
  \caption{Each of the RSL alphabet's 33 signs corresponds to a letter of the Russian language alphabet. Arrows indicate the direction of movement for 8 dynamic signs.}
  \label{fig:gestures}
\end{figure}

\section{Related Work}
\label{sec:related}
This section overviews widespread sign language alphabet datasets. We prioritize isolated alphabetic RSL datasets and cover fingerspelling and isolated datasets in other languages only briefly to focus the reader on the primary topic. Note, however, that datasets of other sign languages can be helpful, for example, for pre-training or developing multilingual sign language models. Since one of this paper's main parts is the data creation process, we also analyze existing methods for collecting and annotating dactyl data, regardless of language.

\subsection{Russian Sign Language Alphabet Datasets}
To our knowledge, only three datasets contain RSL alphabet isolated signs. Two are dedicated to dactyls, while one contains alphabet signs on par with other word signs. \citet{makarov} proposed a dataset of the RSL alphabet divided into 26 static signs, while samples for the remaining 7 dynamic RSL signs were decided not to be collected\footnote{Based on expert consultation, 8 letters are dynamically displayed, but due to uncertainty in the RSL alphabet, the authors of \citet{makarov} consider only 7 letters are shown with movement.}. This choice allowed the authors to create a dataset consisting of only images. 20 people in 20 different scenes recorded 520 samples subsequently cropped by authors. The experiments with this dataset were redundant since it is limited to static images and does not contain dynamic signs. \citet{grif} collected 14,978 samples, images for 25 static signs, and videos for 8 dynamic ones. 4 people participated in the recording in studio conditions. The comparison analysis with this dataset was impossible since it is private. \citet{slovo} presented the dataset Slovo -- the most significant open dataset of isolated RSL signs. Among 20,000 videos, only 660 were provided for 33 dactyl signs, recorded by 30 people. The Slovo was collected on crowdsourcing platforms with significant heterogeneity in recording conditions.

Considered datasets are unsuitable for building a dactyl recognition system limited by conditions in Section~\ref{sec:intro}. Samples from \citet{makarov} were collected only for static signs. Moreover, they are artificially deprived of context using cropping. Also, on par with data from \citet{grif}, the images are homogeneous in subjects, negatively affecting model overfitting. The Slovo dataset's number of samples is insufficient to train a high-quality model. However, the Slovo is an appropriate basis for constructing a more powerful dataset. In contrast to these datasets, Bukva provides (1) the largest number and diversity of signers (155 vs. 4--30), (2) HD/FullHD video samples, and (3) coverage of all 33 RSL alphabet letters, including dynamic ones, in an open-source dataset.

\begin{table*}
\caption{The main parameters of widespread SL alphabets in different languages. The datasets are divided into two groups, containing predominantly images or videos. Relevant Russian SL datasets are highlighted in blue. In the Resolution column, ``crop'' denotes images cropped to the hand region, and ``signals'' denotes data glove sensor recordings rather than RGB samples. The WLASL row refers only to the alphabet-letter subset of this word-level dataset. \newline}
\centering
\scalebox{0.75}{
\begin{tabular}{lccccc}
\hline
Dataset & Letters & Samples & Signers & Resolution & Language\\
\hline\hline
\multicolumn{6}{c}{Images}\\
\hline
MU HandImages ASL, 2011 \citep{new2dstatic} & 26 & 2,425 & 5 & crop & American\\
ASL Fingerspelling, 2015 \citep{realtime} & 24 & 24,000 & 5 & 320 × 240 & American\\
\rowcolor{LightCyan}
RSL dactyl, 2019 \citep{makarov} & 26 & 520 & 20 & crop & Russian\\
BdSL alphabet, 2021 \citep{bangla} & 37 & 35,149 & 350 & crop & Bangla\\
PSL Alphabet 100, 2021 \citep{polish} & 36 & 3,707 & -- & crop & Polish\\
Kazakh Sign Language, 2021 \citep{kz} & 42 & 2,100 & -- & crop & Kazakh\\
\hline
\multicolumn{6}{c}{Videos}\\
\hline
RWTH-Gesture Database, 2006 \citep{rwth} & 30 & 1,200 & 20 & varying & German\\
WLASL, 2020 \citep{li2020WLASL} & 21 & 150 & 12 & varying & American\\
KArSL, 2021 \citep{karsl} & 39 & 5,850 & 3 &  512 × 424 / FullHD & Arabic\\
\rowcolor{LightCyan}
RSL alphabet (NSTU), 2021 \citep{grif} & 33 & 14,978 & 4 & varying & Russian\\
PSL Alphabet, 2023 \citep{psl2023} & 36 & 43,200 & 16 & signals & Polish\\
AzSL dactyl alphabet, 2023 \citep{HASANOV2023102960} & 32 & 13,440 & 221 & varying & Azerbaijani\\
KSU-SSL, 2023 \citep{ksu} & 37 & 18,315 & 33 & varying & Saudi\\
\rowcolor{LightCyan}
Slovo, 2023 \citep{slovo} & 33 & 660 & 30 &  HD / FullHD & Russian\\
NSL23, 2024 \citep{nsl23} & 49 & 1,205 & 23 & varying & Nepali\\
\rowcolor{lemonchiffon}
Bukva, 2026 (ours) & 33 & 3,757 & 155 & HD / FullHD & Russian\\
\hline
\end{tabular}}
\label{tabl:related}
\end{table*}

\subsection{Other Sign Language Alphabet Datasets}
There are multiple dactyl datasets consisting only of static signs. The American SL (ASL) dataset \citep{realtime} includes 24,000 depth maps divided into solely 24 static signs, despite dynamic signs in ASL. Analogously, static signs entirely fill the Japanese SL dataset \citep{jsl} and the Bengali alphabet dataset \citep{bangla}. Another group of dactyl datasets represents dynamic signs with images. MU HandImages ASL \citep{new2dstatic}, the Polish \citep{polish}, and Kazakh \citep{kz} dactylic datasets consist of images for static and dynamic signs. In these datasets, dynamic signs are represented by an image of the hand's final position. The most appropriate choice of modality for dactyl data is video. Such datasets as the RWTH-Gesture Database \citep{rwth}, the KArSL Arabic SL dataset \citep{karsl}, and the KSU-SSL dataset for the Saudi language \citep{ksu} contain video samples for all signs. This subsection also considers fingerspelling datasets to show the difference between continuous and isolated dactyl tasks. The dataset from \citet{lexicon} consists of 3,684 sign sequences of 21,453 letters. The ChicagoFSWild dataset \citep{asl} and its extension \citep{asl_ext} contain 62,536 sign sequences and 302,806 letters. The largest continuous fingerspelling dataset is provided by Google -- American Sign Language Fingerspelling Recognition competition \citep{kaggleasl}; note that it differs in modality, as its samples are hand keypoints rather than videos.

\subsection{Data Collection Pipeline}
Considered datasets can be categorized based on (1) samples' recording method, (2) sign bucket choice, (3) expert choice, (4) additional modalities, and (5) annotation method.

\paragraph{Recording method} One commonly used method for gathering dactyl samples is through studio recording. The dataset from \citet{new2dstatic} was recorded in a studio with a green background and with specially designed moving light sources to create varied lighting effects. The authors of the papers \citep{realtime, lexicon, rwth} also selected a studio recording. Some papers utilized the attempts to diversify the data recorded in studios. So, the KSU-SSL videos \citep{ksu} were recorded by simultaneously employing three cameras, while in \citet{jsl}, the recorders rotated and changed poses during the sign display. The background color was changed while recording the NSL23 dataset \citep{nsl23}. The combination of studio recording and internet downloads is also widespread in the sign video collection. \citet{grif} constructed a dataset of 1,566 samples from YouTube and 13,412 samples recorded in the studio. \citet{asl} proposed a dataset consisting entirely of YouTube videos, while its extended version \citep{asl_ext} was supplemented with videos recorded in the studio. The authors of several datasets received samples through crowdsourcing. Thus, in \citet{makarov}, part of the data was recorded by 15 people in real conditions and transferred to the authors online. Videos of dactyl signs for the Slovo dataset \citep{slovo} were obtained from 30 people through a crowdsourcing platform. The authors validated the data for errors in showing signs due to the perception of crowdworkers as low-skilled.

\paragraph{Sign bucket choice} Depending on whether the selected signs were dynamic or not, the authors decided on the modality of the final dataset: (1) video for dynamic and images for static, (2) video for all signs (if there are dynamic ones), (3) images for all signs (if there are no dynamic ones). However, there are two exceptions: (1) the authors from \citet{new2dstatic} collected images for all dactyl signs of the ASL, including a number for dynamic ones; (2) the authors from \citet{makarov} and \citet{realtime} collected samples only for static signs, ignoring dynamic ones to simplify the data collection process.

\paragraph{Expert choice} Since recording sign videos requires specialized knowledge, selecting signers is integral to creating SL data. The authors of the papers \citep{lexicon, nsl23} involved native SL speakers and beginners just learning it. Videos from the KSU-SSL dataset \citep{ksu} were obtained from individuals specially trained in sign language, followed by expert moderation. The Slovo dataset authors \citep{slovo} restricted video recording task access to signers who had not passed the RSL exam. The exam was structured to only be passable by native speakers and students of the RSL.

\paragraph{Additional modalities} Numerous datasets \citep{nsl23, makarov, kz, jsl, polish, bangla, rwth, slovo} represent images or videos showing hands performing a sign. However, some authors used specialized devices such as Microsoft Kinect V2 \citep{karsl} and Creative Senz3D \citep{realtime} to collect depth maps for each sign. Also, Microsoft Kinect V2 allowed the authors \citep{karsl} to obtain human skeleton annotations. The authors from \citet{jsl} made 3D sign models to receive 8,000 synthetic images of signs from different angles.

\paragraph{Annotation method} Considered datasets practically avoid additional annotation since samples contain a single sign. However, the Slovo \citep{slovo} consists of extra videos that include normal human movement. Such ``sign free'' sequences can further assist the model in detecting signs in a sequence of frames.

\section{Dataset Creation Pipeline}
\label{sec:creation}

We collected videos for all 33 RSL letters via the ABC Elementary and Yandex Toloka crowdsourcing platforms. Following \citet{slovo}, the pipeline comprises three stages: (1) video collection, (2) video filtering and validation, and (3) sign time interval annotation. All crowdworkers passed a mandatory exam to verify RSL proficiency: scoring 80\% or higher granted access to video validation, and 90\% or higher to video recording.

Of approximately 8,400 collected videos, we filtered out low-quality samples and duplicates (8.6\% in total), normalized the rest to 30 FPS, and approved about 6,000 (71.4\%) in validation by at least three workers per video. Three annotators then marked the sign-showing interval in each video. Finally, we combined the approved samples with 660 videos from the Slovo dataset, ensuring at least 101 unique samples per class, and discarded videos where the hand left the frame. See Appendix~\ref{sec:dataset_pipeline} for details, including compensation, exam examples, and the validation and annotation interfaces.

\section{Dataset Characteristics}
\label{sec:dataset}

\begin{figure*}[h]
  \centering
  \includegraphics[width=0.8\linewidth]{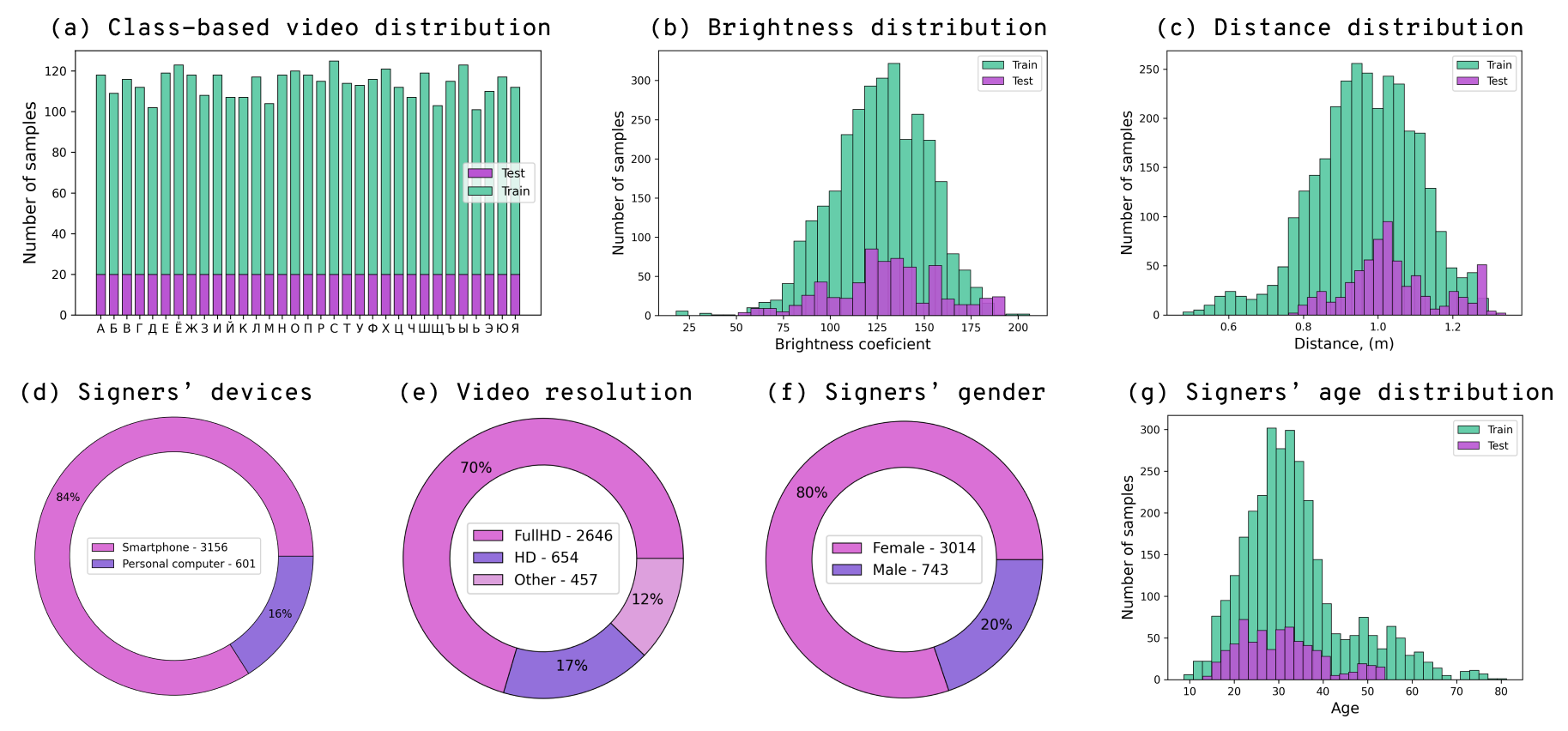}
  \caption{Video resolution, brightness, signers, classes, and splitting analysis. a) class-based video distribution: number of videos for each RSL alphabet letter in the train and test sets; b) brightness distribution: the central image of each video sign part is extracted, converted to grayscale, and then the average pixel brightness is calculated; c) distance distribution: the distance is approximately estimated in meters by computing the length between the signers' left and right shoulders using MediaPipe \citep{mediapipe}; d) signers' devices; e) video resolution; f) signers' gender; g) signers' age distribution: the age is determined by the MiVOLO model \citep{mivolo}.}
  \label{fig:chars}
\end{figure*}

\paragraph{Dataset Versions} We build two versions of the Bukva: an untrimmed version and a trimmed version with an extra 34th class ``no sign'' with 23 GB and 3.2 GB sizes, respectively. 105 additional samples were randomly obtained by annotations from Section~\ref{subsec:annotation} to expand the second set. There is a further description of the initial untrimmed version for clarity. The explicit ``no sign'' class, together with the sign time-interval annotations, additionally supports sign-presence prediction and the temporal localization of fingerspelling within continuous signing. Detailed statistics for each class label are provided in Appendix~\ref{sec:dataset_statistics}.

\paragraph{Dataset Content} The Bukva dataset is a collection of 3,757 videos representing the RSL alphabet. It is divided into 33 signs, each corresponding to a letter of the Russian alphabet. On average, 113 videos per class with an average length of 49 frames were recorded by 155 signers with RSL knowledge. Some participants communicate with sign language daily due to hearing limitations, while others are RSL learners or teachers. The dataset presents videos showing signers aged from $\sim$9 to $\sim$81. The information about including videos with children in the dataset can be found in Section~\ref{sec:ethics}. Most of the videos (80\%) were recorded by women, presumably because most RSL teachers and translators are women. The dataset was primarily gathered indoors and included various scenes and lighting conditions. Figure~\ref{fig:chars} shows the distance to the camera and the video's brightness distributions. Because we release per-signer identifiers together with demographic (age, gender) and acquisition metadata (device, resolution, distance, brightness), Bukva also enables fairness and robustness analyses across signer subgroups and recording conditions, which we encourage the community to explore.

\paragraph{Dataset Size and Quality} About 70\% of the Bukva videos are in FullHD format. Signers recorded the videos using personal devices such as smartphones and laptops (see Figure~\ref{fig:chars}). Since different devices record videos with varying FPS, we standardized all videos to a consistent FPS. 81\% of the videos are in portrait orientation, 16\% are in landscape orientation, and 3\% are in square format. The total duration of the dataset is approximately 4 hours.

\paragraph{Dataset Splitting} The Bukva was divided into train and test sets, comprising 3,097 (82\%) and 660 (18\%) videos, respectively. We intentionally designed the sets to ensure no overlap between the signer groups. Twenty signers who recorded videos for all 33 signs were selected to construct the test set. The train part contains 81 to 105 videos for each sign.

\section{Experiments}
\label{sec:experiments}

Since we are dealing with static and dynamic signs recorded as videos, it is necessary to find a solution to balance image and video classification. Model capacity for image classification is insufficient to cope with dynamics, while 3D architectures are redundant for statics. The TSM block \citep{lin2019tsm} effectively handles this task by modifying the 2D convolutions operation without changing the model architecture and increasing the parameters. Such features enable us to use a pre-trained model with weights from widespread image classification models for the dactyl recognition task. We apply the TSM block to train three architectures: (1) the lightest and fastest MobileNetV2 \citep{sandler2019mobilenetv2}, (2) MobileOne \citep{vasu2023mobileone} from S0 to S2 for a balance between speed and number of parameters, and (3) two ResNet versions \citep{he2015deep} with 18 and 50 blocks as the most widespread image encoders.

\subsection{Experiments Setup}
An essential aspect of training video encoders is a specific strategy for selecting frames from videos. We sample 8 frames uniformly with an individual step calculated by the formula \(step=round(\frac{V_{len}}{N})\), where \(V_{len}\) is the video length, \(N\) is the frame number for sampling, the result is rounded down. This strategy allows us to effectively train models with weak receptive fields since each sampled frame contains a certain proportion of dynamic signs.

We utilize a unified setup for all architectures to facilitate transparent comparisons. An SGD optimizer with an initial learning rate of 0.02, momentum of 0.9, and weight decay of 0.00002 is fixed in the setup. The learning rate is modified by a LinearLR scheduler up to the 20th epoch and then by CosineAnnealingLR \citep{loshchilov2017sgdr} until the end of training. The models are trained on two Tesla V100 with 32GB VRAM, using a batch size of 16 for each card. It is sufficient for all models to converge in 80 epochs.

\subsection{Dataset Capacity Ablation}
Reducing the training set size from 3,182 to 2,000 samples undeniably resulted in an apparent decline in top-1 accuracy, from 0.8368 to 0.7956. A further reduction to 1,000 samples resulted in a significant drop to 0.5676. The number of signers was kept consistent across all three experiments. Similarly, reducing the number of unique signers from 70 to 50 with consistent 1,500 samples decreases the top-1 accuracy from 0.7588 to 0.7415. These results prove an unquestionable decline in top-1 when either the number of samples per class or the diversity of users is reduced. Also, this emphasizes the significance of having a balanced dataset to guarantee strong model performance.

\subsection{Data Preprocessing}
The experiments are based on the trimmed Bukva version with an extra ``no sign'' class, allowing the model to learn to separate signs from other hand movements. Eighty-five samples expand the training set for an additional class, while the test set is expanded by 20, resulting in 3,182 and 680 videos, respectively.

\begin{table*}[tb]
\caption{The models' training results on the Bukva. We evaluate models with top-1 accuracy as the standard metric in the action recognition task. All models are tested on a MacBook M1 Pro using a single core. We measured the model's inference 100 times with further averaging. \newline}
\centering
\scalebox{0.85}{
\begin{tabular}{lcccc}
\hline
\multirow{1}{*}{Model} &
\multirow{1}{*}{Model Size (MB)} &
\multirow{1}{*}{Parameters (M)} &
\multirow{1}{*}{Inference Time (ms)} &
\multirow{1}{*}{top-1 acc.} \\
\hline
ResNet-18 & 44.8 & 11.2 & 191 & 72.7 \\
ResNet-50 & 94.6 & 23.6 & 333 & 79.2 \\
MobileNetV2 & 10.3 & 2.3 & 145 &  \textbf{83.6} \\
MobileOne S0 & 4.4 & 1.1 & 94 &  71.3 \\
MobileOne S1 & 14.1 & 3.5 & 165 &  75.7 \\
MobileOne S2 & 23.4 & 5.8 & 232 &  77.2  \\
\hline
\end{tabular}}
\label{tabl:results}
\end{table*}

\textbf{Training Set.} The vertical and horizontal videos are required to be squared for model input. Primarily, the source video frames are resized to the smaller side of 300 to preserve the aspect ratio. Further, we randomly employ either multi-scale cropping or padding instead of straightforward square resizing to prevent unrealistic spatial stretching of the video. Since dactyl is invariant to the hand side, we reflected the frames vertically by flip augmentation with a probability of 0.5. Multi-scale cropping diversifies the training set regarding hand size and position in the frame, while the flip balances videos on hand orientation. \textbf{Test Set.} Each frame is resized to 224 on the smaller side and padded with 114 to square. We processed the test set without cropping to maintain the dataset's original distribution.

\subsection{Results}
\begin{figure}[h]
  \centering
  \includegraphics[width=1\linewidth]{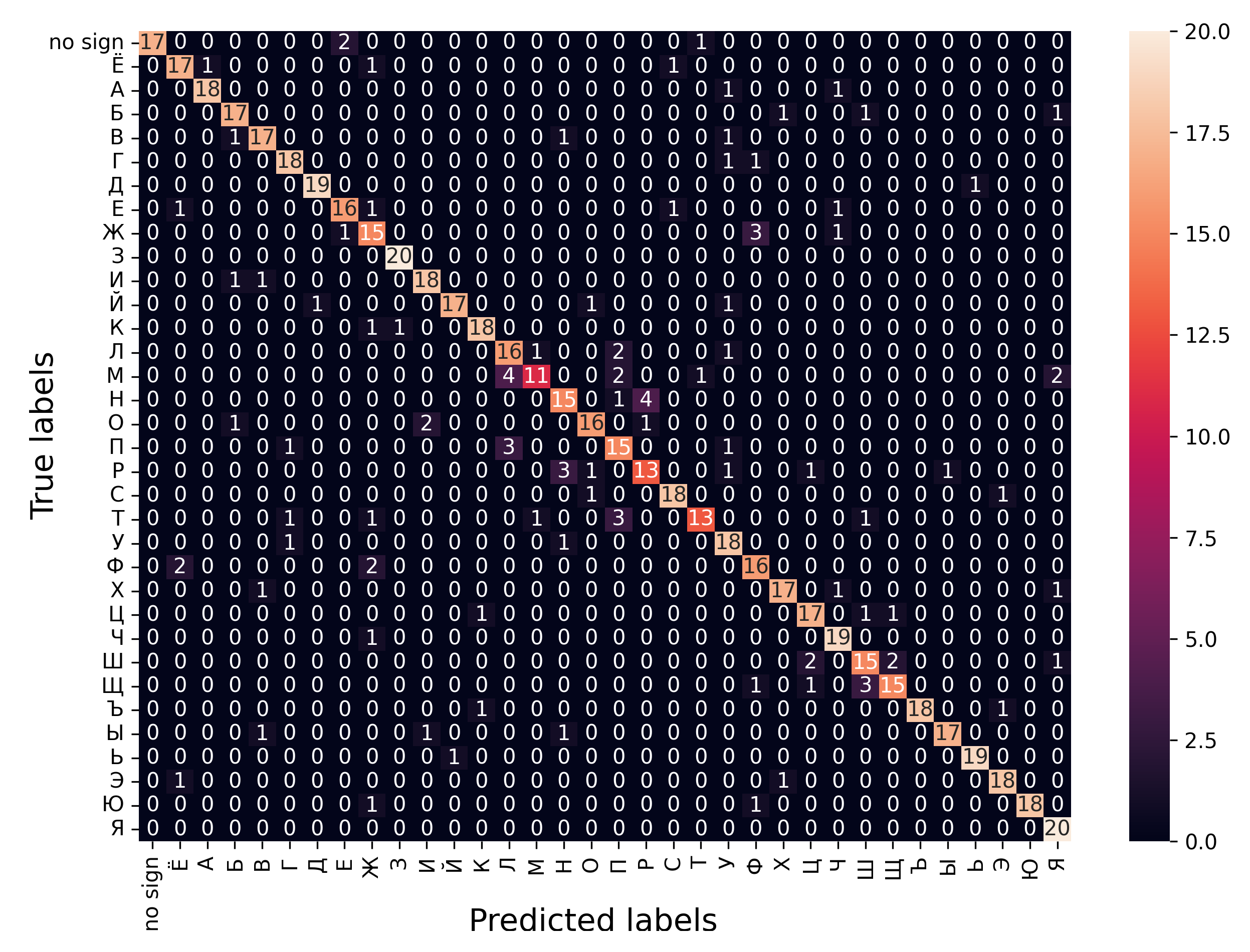}
  \caption{The confusion matrix for test samples using the MobileNetV2 model. Off-diagonal cells indicate misclassifications. The model confuses the classes \foreignlanguage{russian}{``м''} (``m'') and \foreignlanguage{russian}{``л''} (``l''), as their signs are nearly identical.}
  \label{fig:conf_matrix}
\end{figure}

Table~\ref{tabl:results} presents the results of evaluating models using the top-1 accuracy metric. We also provide information about the model sizes, the number of parameters, and the inference time to distinguish the optimal trade-off between quality, model size, and prediction speed. Note that the test set contains only 20 videos per class, so the limited amount of test data results in high error costs and could lead to overestimating model performance. Since ResNet architecture is high-capacity, the small amount of training data and 76\% static signs in the RSL alphabet contributed to model overfitting. Meanwhile, MobileNet performs better due to its relatively compact size. There are multiple false positives in the predictions for similar static signs (see Figure~\ref{fig:conf_matrix}). For example, the letters \foreignlanguage{russian}{``н''} (``n'') and \foreignlanguage{russian}{``р''} (``r'') appear almost identically (see Figure~\ref{fig:gestures}), so the model is frequently mistaken while predicting each of them. The confusion matrix shows that the model struggles to differentiate visually similar signs: errors occur between the dynamic sign \foreignlanguage{russian}{``щ''} (``shch'') and the static sign \foreignlanguage{russian}{``ш''} (``sh''), which have similar finger and hand positions, as well as between signs with similar shapes, such as \foreignlanguage{russian}{``п''} (``p''), \foreignlanguage{russian}{``т''} (``t''), \foreignlanguage{russian}{``м''} (``m''), and \foreignlanguage{russian}{``л''} (``l''). At the same time, the model reliably distinguishes visually contrasting signs. These results suggest that improvements should target sensitivity to subtle differences in hand shapes and movements.

\subsection{Application}
\begin{figure}[h]
  \centering
  \includegraphics[width=0.9\linewidth]{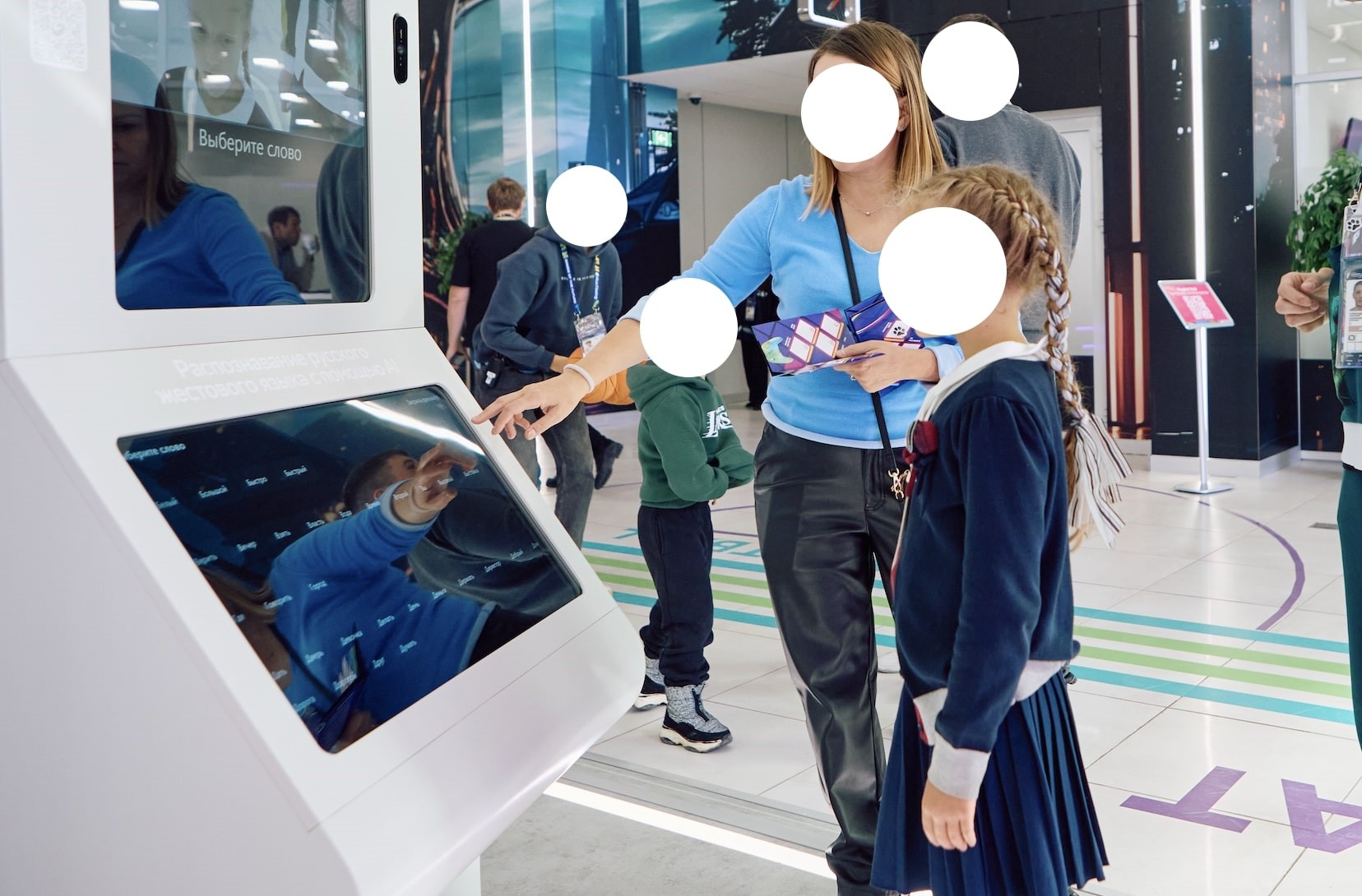}
  \caption{Russian sign language demo stand for education.}
  \label{fig:demo}
\end{figure}

Figure~\ref{fig:demo} shows the demo stand that provides real-time sign language recognition to learn RSL. The user can select any sign from the dataset and learn it offline after watching the corresponding video template. The trained classifier is integrated into the trainer as follows: (1) the user is presented with the correct version of a sign and prompted to replicate it; (2) the system recognizes the sign produced by the user; and (3) it provides a confidence score, which reflects the degree of accuracy in the user's execution of the sign. Thus, a learner who produces a sign slightly incorrectly receives a low confidence score as feedback rather than a simple acceptance of the closest class. A similar approach is used in the browser-based fingerspelling.xyz application for the ASL alphabet. An automatic system for learning any language cannot fully replace a teacher; however, due to the lack of available resources for studying RSL, a dactyl simulator can be a good starting point for building a comprehensive learning system.

\section{Conclusion}
\label{sec:conclusion}
This paper presented Bukva, the first full-fledged open-source video dataset for isolated Russian Sign Language dactyl recognition. The dataset contains 3,757 HD and FullHD videos covering all 33 RSL alphabet letters, including dynamic ones, recorded by 155 signers -- the largest number and diversity of signers among existing RSL alphabet datasets. The crowdsourcing-based creation pipeline with a mandatory RSL exam, multi-stage filtering, validation, and time interval annotation ensured the quality of the data. Baseline experiments with the TSM block demonstrated that a lightweight MobileNetV2 model achieves 83.6\% top-1 accuracy with real-time CPU inference, while the ablation study confirmed the importance of both the number of samples and the diversity of signers. We also presented a demo application illustrating how the trained model can be integrated into an RSL alphabet learning simulator. More broadly, Bukva offers a reusable building block for continuous RSL recognition and translation, and a template for community-driven data collection in other underrepresented sign languages. The dataset, demo code, and pre-trained models will be made publicly available upon acceptance.

\section*{Limitations}
\label{sec:limitations}
Although the Bukva contains enough samples for training lightweight MobileNet, more is required to train ResNet architecture without overfitting. Also, the presence of static signs in the majority may cause such degradation of ResNet's metrics. However, since dactyl recognition applications require only real-time inference, the described problem is irrelevant.

Another limitation concerns the applicability of the proposed dataset to continuous fingerspelling recognition. Continuous fingerspelling includes a significant amount of coarticulation: the appearance of each letter depends on the preceding and following letters, and signs are displayed rapidly from various angles without transitions between them. To fully address the fingerspelling problem, a model must be trained to recognize rapid, subtle movements and the influence of neighboring letters on each letter's appearance. Consequently, while a model trained on the Bukva can serve as a baseline, it is more prone to errors when processing words at high speeds. However, it could help identify slow fingerspelling, which is common in early-stage SL learners with less signing fluency. The Bukva is thus intended to be (1) a pre-training source for a fingerspelling recognition model and (2) a training set for the baseline.

Models can produce inaccurate results when multiple people appear in the same frame, as Bukva's videos contain only one person. Finally, collecting a racially and ethnically diverse RSL dataset presents significant challenges, as the majority of deaf participants and RSL interpreters from crowdsourcing platforms do not represent ethnic minorities. As a result, one of the limitations of the presented dataset is the lack of racial heterogeneity.

\section*{Ethical Considerations}
\label{sec:ethics}
All crowdworkers recording the dataset signed informed consent for data processing and publication. We also ensured that the part of the Slovo dataset \citep{slovo} included in Bukva adheres to the described ethical requirements. We do not restrict videos with signers under 18 since parental permission was obtained during the registration, which complies with the Civil Code of the Russian Federation\footnote{\url{https://ihl-databases.icrc.org/en/national-practice/federal-law-no-152-fz-personal-data-2006}}. To preserve contributors' privacy, we employ anonymized user hash IDs in the dataset annotations. We provide the dataset for research purposes only. However, we recognize that it could be misused for malicious purposes, such as identifying individuals or enabling large-scale surveillance.

\paragraph{Positional Statement} Throughout the research, we involved experts from the ``All-Russian Society of the Deaf'' and professional sign language interpreters. The expertise of the ``All-Russian Society of the Deaf'' (VOG) was utilized at every stage of the Bukva dataset creation. We also engaged deaf consultants in the development of training strategies to apply considerations to specific solutions. Additionally, some of our researchers completed formal courses on RSL to enhance their knowledge in this domain.

\bibliography{egbib}

\appendix

\section{Dataset Creation Details}
\label{sec:dataset_pipeline}

Since the RSL alphabet contains dynamic signs for 8 letters and static for others, we collect videos instead of images for all 33 letters to ensure clarity. We base the data creation pipeline on the one from \citet{slovo}. The pipeline consists of 3 main stages: (1) video collection, (2) video filtering and validation, and (3) sign time interval annotation. \citet{bragg2022exploring} has a nearly similar data creation pipeline and has several necessary steps, which include data recording, quality control, dataset review, and qualitative feedback. We utilized ABC Elementary\footnote{\url{https://elementary.activebc.ru}} as a crowdsourcing platform through all stages. Note that we also employ another crowdsourcing platform, Yandex Toloka\footnote{\url{https://platform.toloka.ai/}}, through the first stage to diversify the group of signers and such context characteristics as background and lighting.

\subsection{Exam}
\begin{figure*}[p]
  \centering
  
  \includegraphics[width=0.75\linewidth]{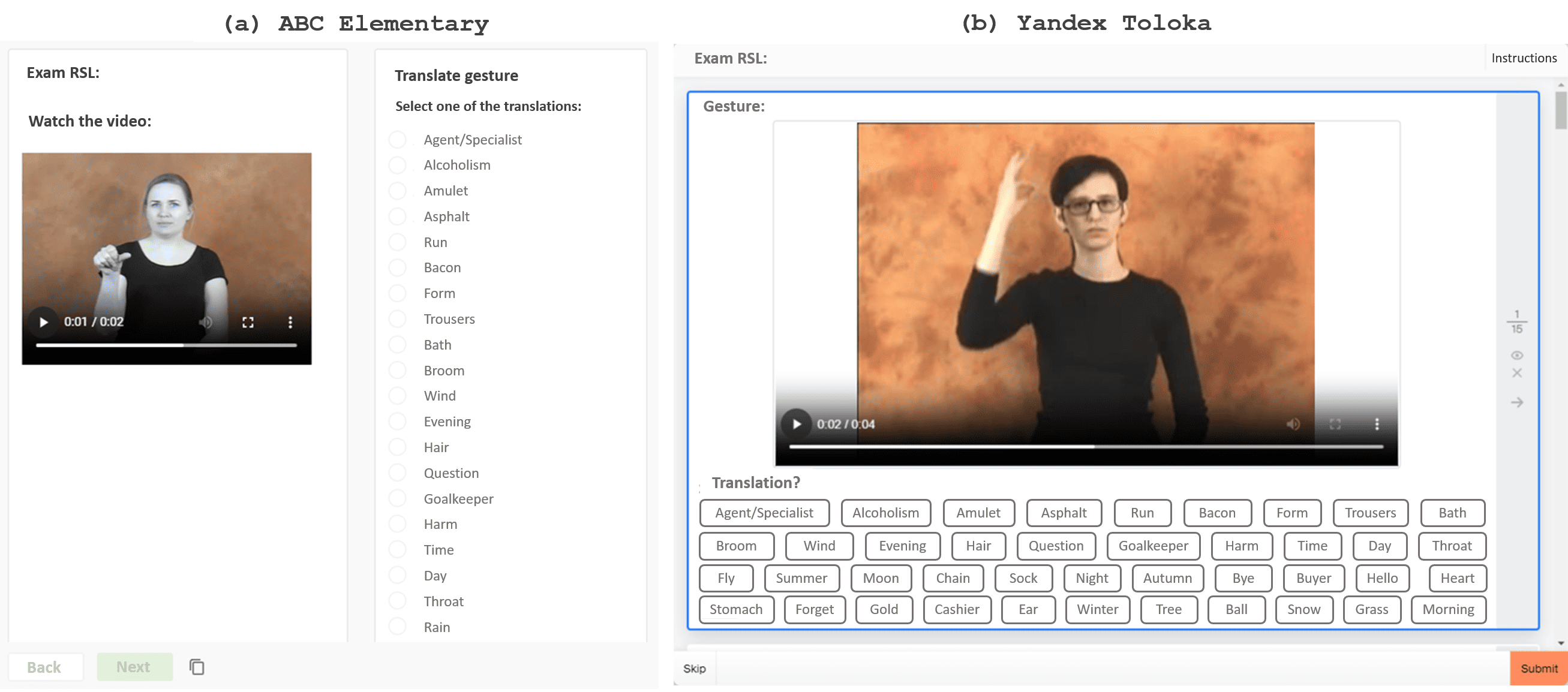}
  \caption{Example of a web page with an exam on crowdsourcing platforms translated into English.}
  \label{fig:exams}
  
  \vspace{1cm}
  
  \includegraphics[width=0.75\linewidth]{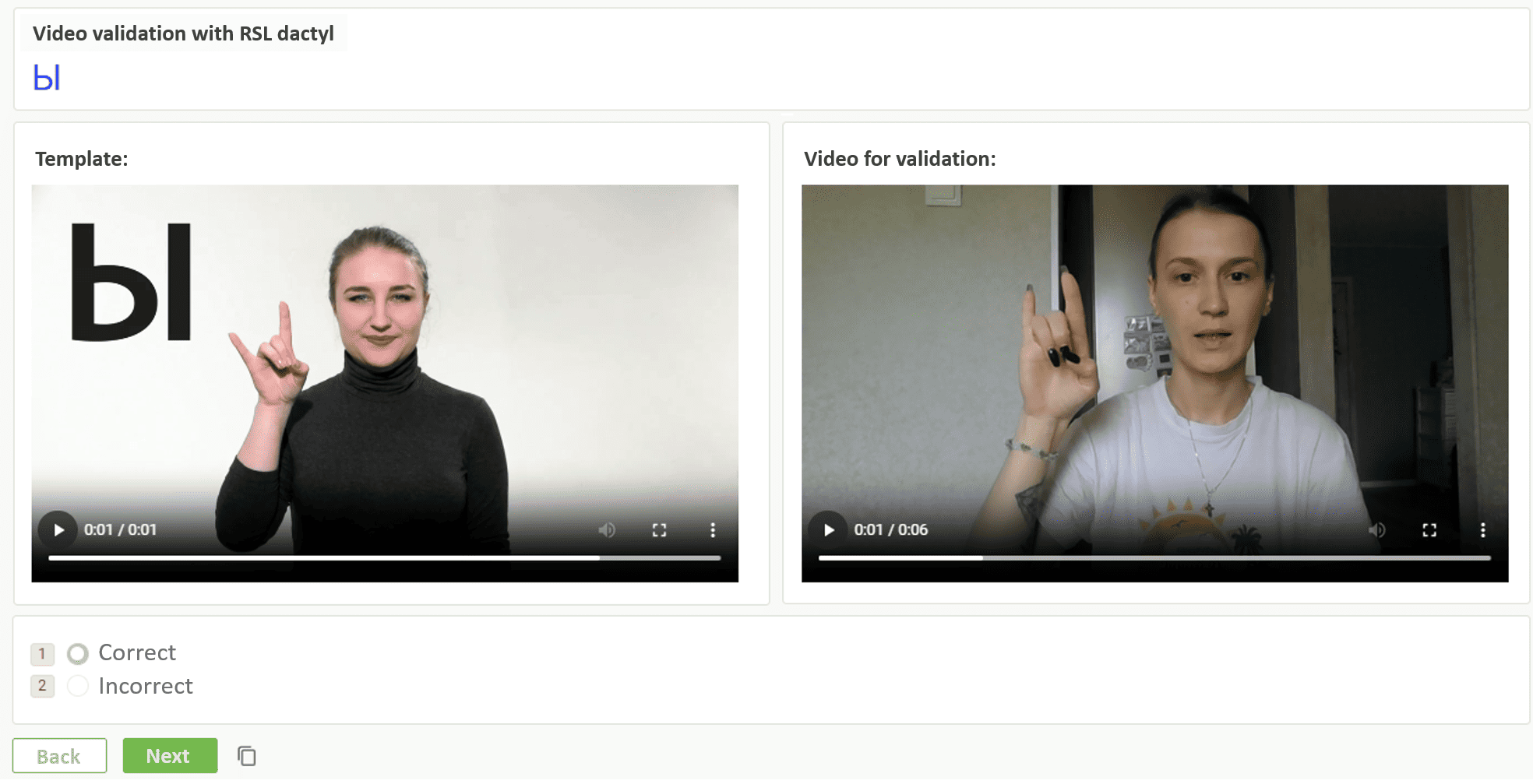}
  \caption{Example of a web page with a video validation on the ABC Elementary crowdsourcing platform translated into English.}
  \label{fig:validation}
  
  \vspace{1cm}
  
  \includegraphics[width=0.75\linewidth]{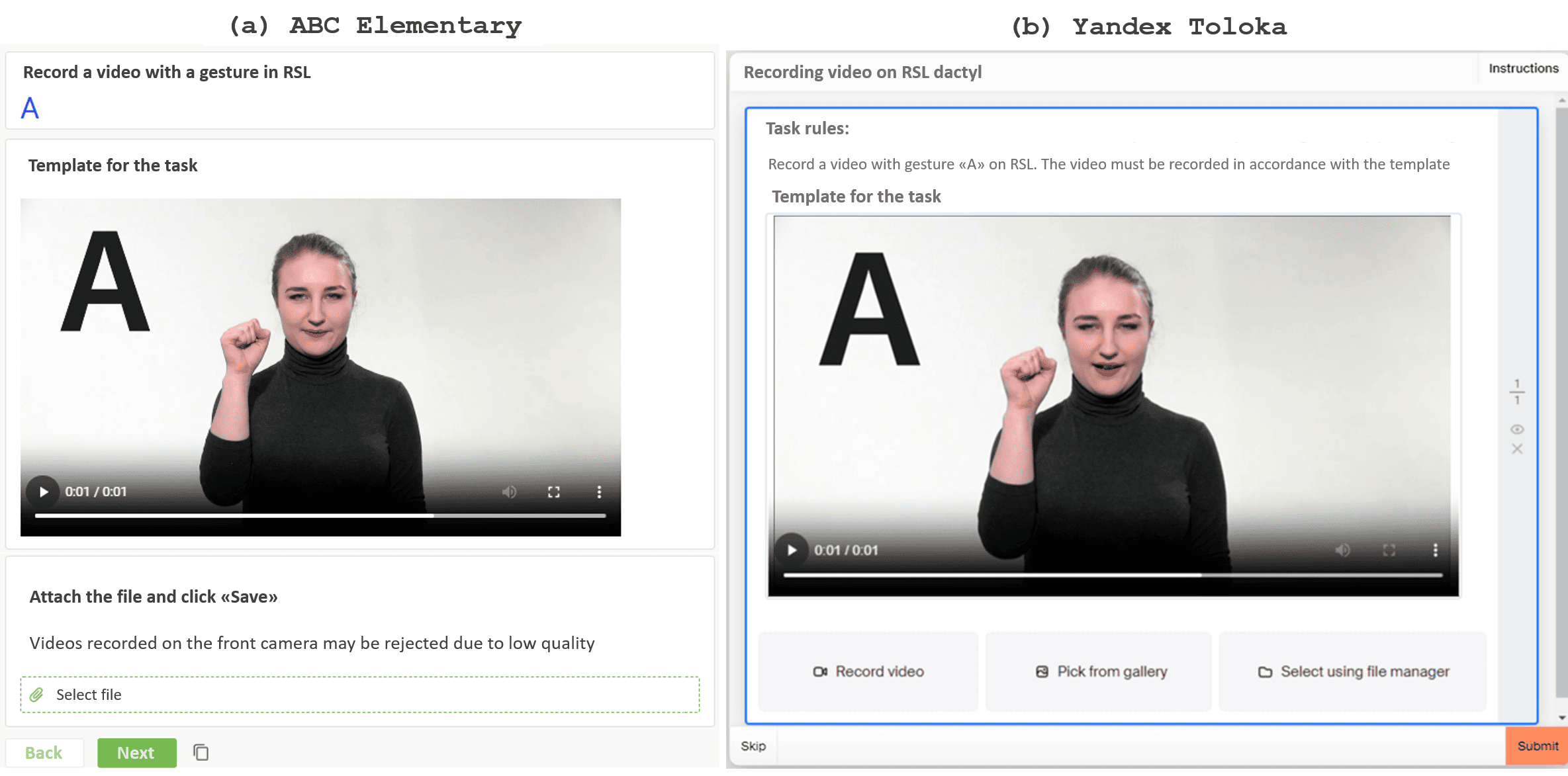}
  \caption{Example of a web page for recording videos on crowdsourcing platforms translated into English.}
  \label{fig:mining}
\end{figure*}

Since Russian sign language knowledge is infrequent, we conducted the qualifying stage to select crowdworkers to show RSL dactyl signs. We asked workers to watch 20 videos from the SpreadTheSign website\footnote{\url{https://www.spreadthesign.com}} demonstrating RSL signs and choose the appropriate translations to spoken Russian language (see Figure~\ref{fig:exams}). The exam time was limited to 10 minutes to avoid fake results from irresponsible people.

\begin{figure*}[p]
  \centering
  \includegraphics[width=0.8\linewidth]{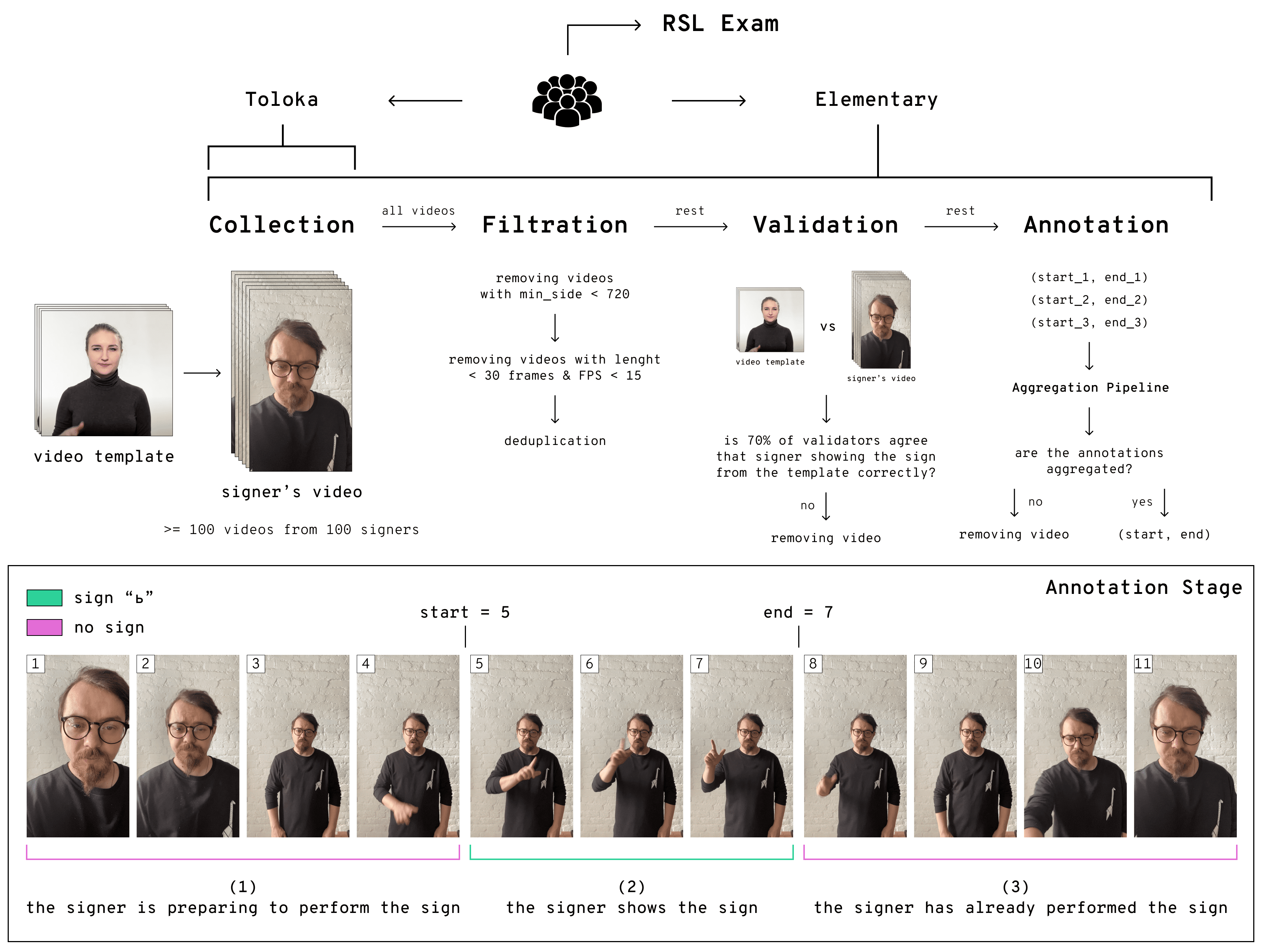}
  \caption{All crowdworkers are required to pass an exam before getting access to all crowd stages. During data collection, signers recorded the sign from the video template with further filtering to remove low-quality videos and duplicates. Only the videos remaining after the filtering stage were verified by at least three crowdworkers to ensure their correctness compared to the template. The annotation stage was also applied only to the samples remaining after the previous stage. Three annotators are asked to mark the video's sign start and end marks. Received marks are aggregated by the algorithm proposed in \citet{slovo}.}
  \label{fig:pipeline}
  
  \vspace{1cm}

  \includegraphics[width=0.75\linewidth]{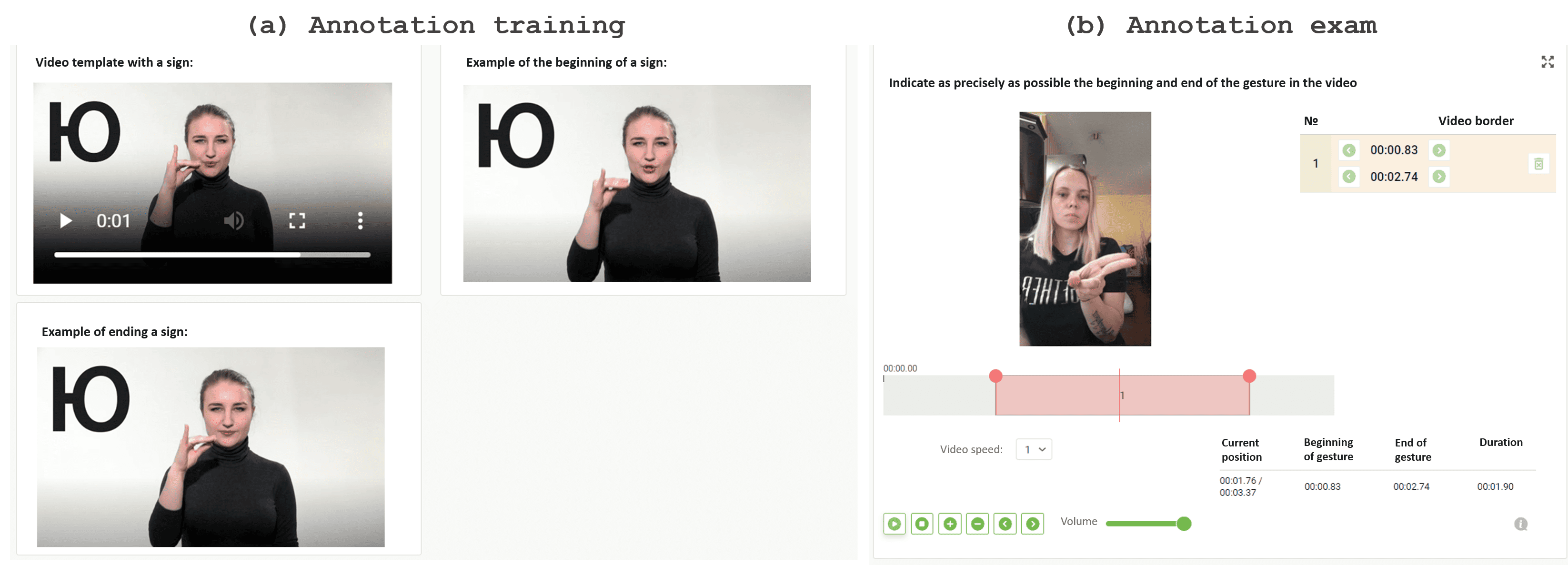}
  \caption{Example of a web page with a video annotation on the ABC crowdsourcing platform translated into English.}
  \label{fig:annotation_abc}
\end{figure*}

Since each stage requires specific RSL alphabet knowledge, the crowdworkers are required to pass a mandatory exam aimed at identifying natives and language learners. Figure~\ref{fig:pipeline} shows the dataset creation process. Hourly wages for assignments were comparable to or slightly higher than the national minimum wage in regions of the Russian Federation. The passed exam was an identifier for workers’ access to video collection and validation stages. Crowdworkers who answered 80\% of exam tasks correctly (at least 16 correct answers) were allowed to validate collected videos (see Figure~\ref{fig:validation}), while access to the video recording stage required at most two mistakes (at least 18 correct answers).

Some participants were involved in crowdsourcing platforms from the ``All-Russian Society of the Deaf'' (VOG), while others were already crowdworkers. The data collection process included a survey on participants' level of deafness and reasons for knowing RSL. The results showed that about 60\% are deaf, 30\% are hard of hearing, and 10\% are professional translators. It was discovered that approximately 20\% of the participants are sign language learners, while the rest are fluent.

\subsection{Video Collection}
\label{subsec:collection}
Crowdworkers were asked to show the RSL dactyl sign specified in the task and record the video through the crowdsourcing platform or load the video from the device's memory. A sign template from the YouTube channel\footnote{\url{https://www.youtube.com/watch?v=jtbwEalS0CE}} was provided for each letter of the RSL alphabet to simplify the process of recording the signs (see Figure~\ref{fig:mining}). We attached instructions to this stage to outline the requirements: (1) only one clothed person is allowed in the frame, (2) the video should be of high quality and remain unshaken throughout, (3) editing or processing the video is prohibited, and (4) a hand performing a sign should be completely visible in the frame. We defined high-quality videos as those with a minimum frame rate of 15 frames per second (FPS), a minimum duration of 30 frames, and a resolution of at least 720 pixels on the smaller side. We have received approximately 8,400 videos at this stage, which will undergo further filtering and validation.

\textbf{Compensation.} We paid \$0.5 for each rule-abiding video, valuing the crowdworker's hour at \$8-10. It is important to note that the average price for common crowdworker tasks is 5 times less. Furthermore, professional sign language translators can earn up to \$10 per hour in this region.

\subsection{Video Filtering and Validation}
\label{subsec:filtration_validation}
\paragraph{Filtering} We left only videos with a frame's short edge of at least 720 pixels to subsequently receive the HD dataset. Also, we filtered out videos shorter than 30 frames and excluded those less than 15 frames per second (FPS). Further, videos were normalized to 30 FPS to align subsequent time-interval annotations. Since crowdsourcing platforms' usage is fraught with unscrupulous workers, the filtering stage was expanded with the deduplication process: crowdworkers may receive a task they have already completed and upload a previously recorded video. We identified duplicates by (1) checking all videos from the same user by user ID and (2) calculating the hash of some video keyframes for the entire dataset. To improve the final dataset's quality, we automatically excluded 286 (3.4\%) low-quality videos. We also identified and rejected 439 (5.2\%) duplicate videos during the deduplication process.

\paragraph{Validation} The validation stage was added to the creation pipeline to check the correctness of sign demonstration. This stage can be considered an extra attempt to avoid wrong sign execution on par with the RSL alphabet exam. The workers were asked to compare the verifiable video with the sign template from the collection stage. Also, validators should check videos for the presence of hands entirely in the frame when a sign is being shown, the absence of other people, traces of processing, and watermarks. Note that including other individuals in the dataset is not feasible due to challenges in obtaining their consent. About 6,000 (71.4\%) samples were approved after this stage.

To increase confidence in the decision, at least three different workers validate each video. In case of disagreement, another crowdworker is required to validate the video, allowing for a maximum of 5 reviewers. We accept samples with at least 70\% positive votes. Note that validators are permitted to check only other people's videos.

Although the validators are not certified RSL instructors, several pipeline properties support the reliability of their judgments. First, only crowdworkers who passed the RSL proficiency exam with at least 80\% correct answers were granted access to this stage. Second, validation is framed as a template-matching task: validators compare the verifiable video against the reference execution of the sign by a professional signer rather than provide a free-form expert judgment of sign correctness. Third, each decision aggregates the votes of three to five independent validators. Finally, as the survey described in Section~\ref{sec:dataset_pipeline} shows, fluent signers, most of whom are deaf or hard of hearing, constitute the majority of the crowdworker pool.

\subsection{Time Interval Annotation}
\label{subsec:annotation}
The process of sign recording frequently includes three parts: (1) the signer is preparing to perform the sign, (2) the signer shows the sign, and (3) the signer has already performed the sign (see Figure~\ref{fig:pipeline}). Since only the second part is informative for the SLR model, the annotation stage was applied to separate the sign frames from the others. We asked the crowdworkers to indicate the time interval on the video with a sign by the sign's start and end marks (see Figure~\ref{fig:annotation_abc}). Each video was annotated by three different workers with subsequent interval aggregation. We utilized the aggregation pipeline from \citet{slovo} without modifications. Furthermore, videos were trimmed along the received marks.

\subsection{Combination and Extra Filtering}
We combined collected samples with 660 dactyl videos from the Slovo dataset to diversify classes on signers. The merged version was cleared of duplicates within a class. We left unique signers per class, receiving at least 101 samples in each.

Although validators checked videos for complete hand presence in the frame, some videos with the hand out of the frame leaked to the annotation stage. We added extra filtering to exclude such samples from the dataset completely. With a hand detector, we receive the bounding box annotations for hands in each video frame. Note that only the sign part\footnote{Those, where the person shows the sign. At number 2 in Section~\ref{subsec:annotation}} of the video is considered. We agreed that the hand is located out of the frame if there are 5 consecutive frames where more than half of the bounding box is outside the frame. The YOLOv7 \citep{yolo7} trained on the HaGRID \citep{hagrid} dataset was utilized as a hand detector.

\section{Dataset Statistics}
\label{sec:dataset_statistics}

\begin{table*}[p]
\caption{Dataset statistics. ``Sign length'' in frames, ``Brightness'' in grayscale levels, ``Gender'', ``Devices'' by records, ``Age'' of crowdworkers and ``Total'' number of videos. The Latin transliteration of each Cyrillic letter is given in brackets. \newline}
\centering
\scalebox{0.75}{
\begin{tabular}{|c|ccc|ccc|cc|cc|ccc|c|}
\hline
\textbf{Letters} & \multicolumn{3}{c|}{\textbf{Sign length}} & \multicolumn{3}{c|}{\textbf{Brightness}} & \multicolumn{2}{c|}{\textbf{Gender}} & \multicolumn{2}{c|}{\textbf{Devices}} & \multicolumn{3}{c|}{\textbf{Age}} & \textbf{Total} \\
\cline{2-14}
& Min & Max & Median & Min & Max & Median & Male & Female & Smartphone & PC & Min & Max & Median & \\
\hline
\foreignlanguage{russian}{А} (A) & 23 & 94 & 45.0 & 59 & 185 & 131 & 26 & 92 & 101 & 17 & 11.1 & 71.3 & 30.7 & 118\\
\foreignlanguage{russian}{Б} (B) & 21 & 112 & 50.0 & 62 & 191 & 132 & 17 & 92 & 94 & 15 & 13.5 & 76.6 & 31.7 & 109\\
\foreignlanguage{russian}{В} (V) & 22 & 105 & 44.0 & 57 & 192 & 129 & 21 & 95 & 95 & 21 & 10.4 & 71.7 & 32.4 & 116\\
\foreignlanguage{russian}{Г} (G) & 25 & 100 & 43.0 & 69 & 190 & 125 & 19 & 92 & 94 & 18 & 12.7 & 63.9 & 31.4 & 112\\
\foreignlanguage{russian}{Д} (D) & 32 & 105 & 50.0 & 65 & 190 & 130 & 20 & 82 & 84 & 18 & 11.4 & 66.1 & 31.1 & 102\\
\foreignlanguage{russian}{Е} (E) & 22 & 101 & 47.0 & 66 & 186 & 129 & 22 & 97 & 99 & 20 & 10.8 & 74.4 & 31.8 & 119\\
\foreignlanguage{russian}{Ё} (Yo) & 27 & 132 & 54.0 & 78 & 184 & 127 & 27 & 96 & 104 & 19 & 10.9 & 73.6 & 31.1 & 123\\
\foreignlanguage{russian}{Ж} (Zh) & 22 & 125 & 48.5 & 31 & 206 & 123 & 28 & 90 & 99 & 19 & 12.0 & 77.6 & 29.6 & 118\\
\foreignlanguage{russian}{З} (Z) & 28 & 92 & 51.0 & 67 & 191 & 127 & 24 & 84 & 92 & 16 & 11.8 & 75.3 & 31.9 & 108\\
\foreignlanguage{russian}{И} (I) & 21 & 138 & 46.0 & 67 & 191 & 129 & 19 & 99 & 101 & 17 & 9.4 & 74.3 & 31.4 & 118\\
\foreignlanguage{russian}{Й} (Y) & 25 & 111 & 48.0 & 65 & 194 & 129 & 22 & 85 & 89 & 18 & 8.7 & 75.0 & 31.7 & 107\\
\foreignlanguage{russian}{К} (K) & 17 & 121 & 43.0 & 19 & 193 & 130 & 20 & 87 & 93 & 14 & 11.1 & 73.4 & 31.1 & 107\\
\foreignlanguage{russian}{Л} (L) & 22 & 123 & 45.0 & 66 & 191 & 125 & 25 & 92 & 97 & 20 & 13.8 & 72.7 & 31.8 & 117\\
\foreignlanguage{russian}{М} (M) & 24 & 140 & 46.5 & 63 & 177 & 125 & 21 & 83 & 89 & 15 & 10.0 & 72.1 & 31.1 & 104\\
\foreignlanguage{russian}{Н} (N) & 28 & 124 & 47.0 & 54 & 189 & 128 & 22 & 96 & 99 & 19 & 14.2 & 65.5 & 31.5 & 118\\
\foreignlanguage{russian}{О} (O) & 26 & 100 & 44.5 & 60 & 191 & 128 & 21 & 99 & 100 & 20 & 16.4 & 75.9 & 31.1 & 120\\
\foreignlanguage{russian}{П} (P) & 22 & 124 & 47.0 & 58 & 190 & 126 & 22 & 96 & 95 & 23 & 10.8 & 71.9 & 33.0 & 118\\
\foreignlanguage{russian}{Р} (R) & 23 & 90 & 50.0 & 60 & 191 & 128 & 22 & 93 & 98 & 17 & 11.7 & 74.2 & 30.6 & 115\\
\foreignlanguage{russian}{С} (S) & 27 & 102 & 44.0 & 18 & 189 & 125 & 27 & 98 & 102 & 23 & 12.8 & 75.2 & 31.7 & 125\\
\foreignlanguage{russian}{Т} (T) & 24 & 104 & 46.0 & 58 & 183 & 126 & 21 & 93 & 94 & 20 & 13.4 & 75.3 & 31.5 & 114\\
\foreignlanguage{russian}{У} (U) & 24 & 182 & 45.0 & 31 & 193 & 129 & 21 & 92 & 98 & 15 & 10.1 & 62.5 & 30.4 & 113\\
\foreignlanguage{russian}{Ф} (F) & 24 & 133 & 48.0 & 59 & 185 & 126 & 22 & 94 & 95 & 21 & 12.2 & 76.4 & 31.1 & 116\\
\foreignlanguage{russian}{Х} (Kh) & 24 & 116 & 45.0 & 67 & 186 & 130 & 25 & 96 & 100 & 21 & 11.9 & 72.7 & 30.9 & 121\\
\foreignlanguage{russian}{Ц} (Ts) & 24 & 85 & 47.0 & 51 & 200 & 128 & 24 & 86 & 93 & 19 & 12.5 & 74.6 & 31.8 & 112\\
\foreignlanguage{russian}{Ч} (Ch) & 27 & 104 & 50.0 & 61 & 193 & 127 & 25 & 87 & 87 & 20 & 11.1 & 74.8 & 30.6 & 107\\
\foreignlanguage{russian}{Ш} (Sh) & 28 & 132 & 49.0 & 18 & 194 & 129 & 23 & 96 & 100 & 19 & 12.2 & 73.3 & 30.5 & 119\\
\foreignlanguage{russian}{Щ} (Shch) & 21 & 93 & 46.0 & 24 & 188 & 127 & 21 & 82 & 87 & 16 & 14.6 & 72.4 & 30.1 & 103\\
\foreignlanguage{russian}{Ъ} (hard sign) & 26 & 124 & 48.0 & 55 & 191 & 128 & 26 & 89 & 97 & 18 & 13.9 & 81.2 & 32.2 & 115\\
\foreignlanguage{russian}{Ы} (Y) & 23 & 174 & 43.0 & 73 & 185 & 127 & 24 & 99 & 105 & 18 & 11.7 & 72.9 & 31.9 & 123\\
\foreignlanguage{russian}{Ь} (soft sign) & 24 & 124 & 44.0 & 21 & 187 & 131 & 18 & 83 & 90 & 11 & 12.3 & 64.0 & 31.7 & 101\\
\foreignlanguage{russian}{Э} (E) & 22 & 156 & 47.0 & 22 & 191 & 129 & 24 & 86 & 92 & 18 & 9.8 & 73.6 & 30.9 & 110\\
\foreignlanguage{russian}{Ю} (Yu) & 19 & 129 & 50.0 & 62 & 192 & 126 & 23 & 94 & 99 & 18 & 11.8 & 74.2 & 31.1 & 117\\
\foreignlanguage{russian}{Я} (Ya) & 25 & 139 & 47.5 & 73 & 191 & 128 & 24 & 88 & 94 & 18 & 11.3 & 72.3 & 31.5 & 112\\
\hline
\end{tabular}}
\label{tabl:data}
\end{table*}

\end{document}